\documentclass{article}

\usepackage{microtype}
\usepackage{graphicx}
\usepackage{subcaption}
\usepackage{booktabs} 

\usepackage{hyperref}

\usepackage[preprint]{icml2026}

\usepackage{bm}
\usepackage{amsmath}
\usepackage{amssymb}
\usepackage{mathtools}
\usepackage{amsthm}

\usepackage[capitalize,noabbrev]{cleveref}

\theoremstyle{plain}
\newtheorem{theorem}{Theorem}[section]

\theoremstyle{definition}

\theoremstyle{remark}

\usepackage[textsize=tiny]{todonotes}
\usepackage{multirow}

\icmltitlerunning{Penalizing Localized Dirichlet Energies in Low Rank Tensor Products}

\begin{document}

\twocolumn[
  \icmltitle{Penalizing Localized Dirichlet Energies in Low Rank Tensor Products}



  \icmlsetsymbol{equal}{*}

  \begin{icmlauthorlist}
    \icmlauthor{Paris A. Karakasis}{yyy}
    \icmlauthor{Nicholas D. Sidiropoulos}{yyy}
  \end{icmlauthorlist}

   \icmlaffiliation{yyy}{Department of Electrical and Computer Engineering, University of Virginia, Charlottesville, USA}

  \icmlcorrespondingauthor{Paris Karakasis}{karakasis@virginia.edu}


  \vskip 0.3in
]



\printAffiliationsAndNotice{}  

\begin{abstract}
We study low-rank tensor-product B-spline (TPBS) models for regression tasks and investigate Dirichlet energy as a measure of smoothness. We show that TPBS models admit a closed-form expression for the Dirichlet energy, and reveal scenarios where perfect interpolation is possible with exponentially small Dirichlet energy. This renders global Dirichlet energy-based regularization ineffective. To address this limitation, we propose a novel regularization strategy based on local Dirichlet energies defined on small hypercubes centered at the training points. Leveraging pretrained TPBS models, we also introduce two estimators for inference from incomplete samples. Comparative experiments with neural networks demonstrate that TPBS models outperform neural networks in the overfitting regime for most datasets, and maintain competitive performance otherwise. Overall, TPBS models exhibit greater robustness to overfitting and consistently benefit from regularization, while neural networks are more sensitive to overfitting and less effective in leveraging regularization. 
\end{abstract}

\section{Introduction}

Classification and regression are classical yet fundamentally ill-posed problems in machine learning. 
In regression, for example, given a collection of samples $\mathcal{C} = \{(\mathbf{x}_m, y_m)\}_{m=1}^M$, drawn from an unknown distribution and evaluated through an unknown function, there exist infinitely many functions that perfectly interpolate $\mathcal{C}$. A similar phenomenon arises in classification: for virtually any finite training set, one can construct infinitely many classifiers that achieve zero training error. In both cases, this raises a central question: among all hypotheses consistent with the observed data, which ones generalize best to unseen samples drawn from the same distribution?

Without additional assumptions, this question is essentially vacuous. 
In many practical settings, only scarce samples are available, while more expressive models typically require larger datasets to yield reliable estimates that generalize effectively. 
To address this challenge, a wide range of function-complexity measures have been developed to combat overfitting and encourage the selection of simpler hypotheses that are more likely to generalize. 
The need for additional constraints to render learning problems well-posed can be traced  back to Hadamard's foundational work on ill-posedness \cite{hadamard1902problemes}. 
Building on this principle, Tikhonov formalized explicit regularization methods for inverse problems, introducing penalty terms that promote stability and simplicity \cite{tikhonov1977solutions}. 
These ideas were further developed within statistical learning theory, where Vapnik and others established regularization as a central mechanism for controlling model complexity and ensuring strong generalization \cite{vapnik2013nature}. 
Designing effective regularization principles that bias learning toward hypotheses with good generalization remains an active research direction.

The function approximation model that has attracted the most attention in recent years is deep neural networks (DNNs). 
However, designing effective regularization methods for DNNs remains challenging \cite{gouk2021regularisation}. 
The most widely used techniques, such as dropout \cite{srivastava2014dropout} and batch normalization \cite{ioffe2015batch}, are largely heuristically motivated \cite{gouk2021regularisation}, which complicates the design and improvement of new methods. In contrast, classical approaches such as Tikhonov regularization are generally less effective for deep architectures \cite{srivastava2014dropout}, underscoring the need for principled and scalable regularization strategies. 

Yet another conundrum emerges from the empirical observation that large over-parameterized DNNs often generalize better than their smaller counterparts, which conflicts with classical notions of model parsimony which favor smaller models \cite{novak2018sensitivity}. This paradoxical phenomenon is known as the double-descent \cite{belkin2021fit, nakkiran2021deep, dherin2021geometric} -- see also the appendix of this manuscript where we reproduce this phenomenon using the well-known MNIST dataset and neural networks with only one hidden layer.

Despite extensive research, there is still no strong consensus on what controls generalization in deep neural networks. 
Current efforts largely focus on classification problems, where the targets $\{y_m\}$ are drawn from a finite set. 
In this setting, several factors have been investigated to understand generalization, including Rademacher complexity \cite{shalev2014understanding}, VC dimension \cite{vapnik2013nature}, early stopping \cite{hardt2016train}, batch size effects \cite{keskar2016large}, the magnitudes and norms of the network weights \cite{bartlett1996sample, bartlett2017spectrally, neyshabur2017pac, gouk2021regularisation}, sharpness \cite{keskar2016large}, and the choice of optimization algorithm \cite{neyshabur2017exploring}. 
Researchers have explored these factors both empirically, by observing correlations with generalization, and theoretically, by attempting to explain these correlations.

Following the intuition that slowly varying functions are simpler and generalize better, the Lipschitz constant has emerged as a crucial factor in assessing and enhancing the performance of DNNs \cite{fazlyab2019efficient, gouk2021regularisation, bubeck2021universal}. 
\citet{xu2012robustness} establish a formal link between a model's Lipschitz constant and its generalization performance via robustness theory, and \cite{gouk2021regularisation} provide empirical evidence that enforcing Lipschitz control, e.g., through layer-wise projections, improves generalization. 
\citet{khromov2023some} observe a striking double-descent trend in both upper and lower bounds of the Lipschitz constant during training. Theoretical work also highlights the interplay between model size, smoothness, and generalization. \citet{bubeck2021universal} show that interpolating $n$ points in $\mathbb{R}^D$ with a smooth function requires at least $nD$ parameters, yielding a size--robustness tradeoff: models with only $n$ parameters are necessarily non-robust (large Lipschitz constant), whereas achieving robustness (small Lipschitz constant) requires larger models.

\citet{dherin2021geometric} argue that overparameterized networks trained with stochastic gradient descent are implicitly regularized via \emph{geometric complexity}, which depends on the Dirichlet energy of the function. Gradient-based training induces an implicit preference for functions with low Dirichlet energy, so the learned DNNs often behave like harmonic interpolants of the data rather than arbitrary fits \cite{dherin2021geometric, dherin2022neural}. Moreover, common training heuristics—including parameter norm regularization, spectral norm regularization, flatness regularization, implicit gradient regularization, noise regularization, and parameter initialization—act to control geometric complexity, providing a unifying framework for understanding deep learning \cite{dherin2022neural}. Empirical studies support this view: a complexity measure similar to geometric complexity correlates strongly with generalization in DNNs \cite{novak2018sensitivity}, and recent work derives generalization bounds that scale with the margin-normalized geometric complexity of DNNs \cite{munn2023margin}. 
However, this behavior is not universal across all learning models, as interpolation with functions of exponentially small Dirichlet energy is possible, casting doubt on the reliability of Dirichlet energy as a universal proxy for generalization \cite{bubeck2021universal}.

In this paper, we study low-rank tensor products of univariate functions as an alternative function approximation model. Such models are compatible with calculus operations. This enables the derivation of analytical insights on many interesting questions -- even closed-form solutions in certain cases. We show that the Dirichlet energy of these low-rank tensor product models can be computed exactly in closed form, and that this computation can be carried out efficiently with a cost that scales linearly in the dimensionality of the problem. This provides a clearer understanding of the model’s behavior and explains why these representations can perfectly interpolate data while maintaining exponentially small Dirichlet energy. Motivated by this observation, we propose using localized Dirichlet energies, centered at the training points, as a regularization strategy for both regression and classification tasks.

\section{Background \& Related Work}

In this section, we review the main concepts and tools underlying our proposed framework. We focus on low-rank tensor models for multivariate function approximation with B-spline representations for the univariate components. Additionally, we introduce the Dirichlet energy as a indirect measure of the geometric complexity of functions, following \cite{dherin2021geometric, dherin2022neural}, which we later use to promote smoothness and regularity in our models. 

\subsection{Low Rank Tensor Product models}

Low-rank tensor products of univariate functions are commonly used to approximate multivariate functions $\mathbf{g}(\mathbf{x}): [0,1]^N \rightarrow \mathbb{R}^M$
and can be represented as a sum of \(R\) separable components, i.e.,
\begin{equation}
    \mathbf{g}(\mathbf{x}) = \sum_{r=1}^R \mathbf{v}_r \prod_{n=1}^N g_{n,r}(x_n),
    \label{CPD}
\end{equation}
where \(g_{n,r} : [0,1] \rightarrow \mathbb{R}\) are univariate functions along the \(n\)-th variable, \(\mathbf{v}_r \in \mathbb{R}^{M}\) are vectors associated with each separable component, and \(R\) is typically chosen to be small.

\citet{kargas2021supervised} recently proposed the model in (\ref{CPD}) for supervised learning of smooth multivariate functions. More specifically, they considered compactly supported\footnote{Here we consider functions that are compactly supported on the $[0,1]^N$ hypercube; without loss of generality, any compactly supported function can be mapped to this domain via appropriate translation and rescaling of its arguments.} functions $f: [0, 1]^N \rightarrow \mathbb{R}.$ Their approach leverages a periodic extension of these functions, which allows them to be efficiently approximated using truncated multivariate Fourier series, yielding a finite parametrization of the model. This parametrization naturally appears in the form of an $N$-th order tensor of coefficients. After considering a constrained decomposition of that tensor, \citet{kargas2021supervised} arrive at the model in (\ref{CPD}) for $M=1$, where the univariate functions $g_{n,r}: [0,1] \rightarrow \mathbb{R}$ are modeled using truncated Fourier series.

\citet{govindarajan2022regression} proposed using B-splines, or equivalently piecewise polynomials, to discretize the univariate component functions \(g_{n,r} : [0,1] \rightarrow \mathbb{R}\) in the low-rank tensor model of (\ref{CPD}). A spline is a piecewise polynomial function stitched together to maintain a specified degree of smoothness across its domain. B-splines have long been fundamental tools in computational geometry and numerical analysis \cite{de1978practical}. However, their explicit use in machine learning has been comparatively limited, despite known connections between deep ReLU networks and nonuniform linear splines with adaptive knots \cite{balestriero2018spline, unser2019representer}. Compared to global polynomials, B-splines offer compact support, allow adaptive local approximation of nonlinearities, and avoid the oscillatory behavior often seen in high-degree polynomial interpolation. Furthermore, B-splines are universal approximators for continuous functions on compact intervals \cite{de1978practical}.

Unlike global polynomials, whose approximation power grows only by increasing the degree, splines can also be refined by inserting knots. This improves local resolution and enables precise local adaptation. Changes to coefficients affect only a small neighborhood, allowing accurate modeling of regions with rapidly varying behavior. Low-degree splines avoid numerical instabilities such as the large oscillations of Runge’s phenomenon \cite{trefethen2019approximation, schumaker2007spline}. Such phenomena are known to adversely affect generalization in machine learning. Their compact support also makes splines particularly effective for capturing localized non-analytic features, such as kinks and sharp transitions. Global polynomial and Fourier bases, by contrast, can perform poorly in these situations \cite{powell1981approximation, rivlin1981introduction, daubechies2004iterative}. 

Global polynomial and Fourier bases have complementary strengths. Their global support allows information to propagate across the entire domain, enabling interpolation or reasonable approximation in regions with few or no training samples, provided the underlying function is sufficiently smooth. However, they lack local adaptability and can exhibit oscillatory artifacts near boundaries or discontinuities (Runge and Gibbs phenomena). By contrast, B-splines provide a controlled, locally supported model, which, when combined with explicit regularization, allows information to propagate across the domain in a controllable manner while maintaining stability and precise local control. These properties make low-rank tensor products of univariate B-splines a practical and theoretically sound choice for high-dimensional regression and classification. In this work, we use low rank tensor products of univariate B-splines as our function approximation model.

\subsection{Dirichlet energy as a measure of complexity}

The geometric complexity of a function $\mathbf{g}:[0,1]^N\rightarrow\mathbb{R}^M$ can be intuitively defined as the volume of its graph 
\begin{equation}
    \text{gr}(\mathbf{g}) := \{ (\mathbf{x}, \mathbf{g}(\mathbf{x})) : \mathbf{x} \in [0,1]^N \} \subset \mathbb{R}^{N+M}.
\end{equation}
For smooth functions, this graph is an $N$-dimensional submanifold, and its Riemannian volume, induced by the Euclidean metric, provides a measure of the functions's geometric complexity \cite{do2016differential,dherin2021geometric}
\begin{equation}
    \Omega(\mathbf{g}) = \int_{[0,1]^N} \sqrt{1 + \left\|\nabla \mathbf{g}(\mathbf{x})\right\|^2_F} \, d\mathbf{x}.
\end{equation}
Using the first-order Taylor approximation $\sqrt{1+z} \approx 1 + \frac{1}{2} z$, this reduces to
\begin{equation}
    \Omega(\mathbf{g}) \approx 1 +  \frac{1}{2} \text{DE}(\mathbf{g}),
\end{equation}
where
\begin{equation}
    \text{DE}(\mathbf{g}) := \int_{[0,1]^N} \left\|\nabla \mathbf{g}(\mathbf{x})\right\|_F^2 \, d\mathbf{x}
\end{equation}
is the Dirichlet energy of $\mathbf{g}$. Thus, the Dirichlet energy provides a simple and effective proxy for the geometric complexity of a function.

This notion of geometric complexity has a direct connection to deep learning: the Jacobian of a neural network captures how sensitive the network output is to input perturbations, which is directly related to the Dirichlet energy. Computing the exact Dirichlet energy of a high-dimensional DNN is intractable, as it requires integrating over the entire input space. However, sample-based approximations are practical. Evaluating $\left\|\nabla \mathbf{g}(\mathbf{x})\right\|^2_F$ at the training points and averaging provides an empirical estimate of the Dirichlet energy. In particular, averaging the Frobenius norm of the Jacobian over the training set has been proposed as an effective regularizer to improve robustness and generalization \cite{sokolic2017robust, hoffman2019robust}. Furthermore, empirical and theoretical studies further support that geometric complexity correlates strongly with generalization in DNNs \cite{novak2018sensitivity, munn2023margin}.

\section{Main Contributions}

In this section, we present three main contributions. First, we show that the Dirichlet energy of low-rank tensor product models can be computed analytically in closed form. This result allows us to gain insights into how near-perfect interpolations of datasets with exponentially small Dirichlet energy are possible. Second, we propose a regularization method based on localized Dirichlet energies for supervised learning tasks, including classification and regression. Third, we introduce a framework for inference with incomplete data, i.e., sample vectors $\mathbf{x}$ with incomplete elements. These contributions are agnostic to the choice of univariate function representations in low-rank tensor-product models, such as Fourier series or B-splines. Our framework distinguishes itself from prior approaches, including \cite{kargas2021supervised} and \cite{govindarajan2022regression}, in three key aspects: (i) we introduce averaged localized Dirichlet energy as a regularization term for learning, (ii) we propose two marginalization strategies for run-time inference in the presence of missing data, and (iii) we place particular emphasis on the overfitting regime, providing empirical evidence for the behavior of the proposed model and its connection to the double-descent phenomenon.

\subsection{The Dirichlet energy of a low rank tensor model}

One advantage of the low-rank tensor product models in (\ref{CPD}) is their compatibility with basic calculus operations. In particular, this structure allows us to analytically compute the Dirichlet energy in closed form, providing insight into the model's behavior and inner mechanisms.

\begin{theorem}
For a function $\mathbf{g}$ represented by the low-rank tensor product model in (\ref{CPD}), its Dirichlet energy can be expressed in closed form as
\begin{equation}
\text{DE}(\mathbf{g}) = \int_{[0,1]^N} \left\|\nabla \mathbf{g}(\mathbf{x})\right\|_F^2 \, d\mathbf{x}
= \mathbf{s}(\mathbf{g})^\top \mathbf{Z}(\mathbf{g}) \mathbf{s}(\mathbf{g}),
\label{Theorem1a}
\end{equation}
where $\mathbf{s}(\mathbf{g}) \in \mathbb{R}^R_+$, 
$\mathbf{Z}_{r,s}(\mathbf{g}) \in \mathbb{R}^{R \times R}$, 
$\mathbf{A}^{(0)}(\mathbf{g}) \in \mathbb{R}^{R \times R}$, and $\mathbf{A}^{(1)}(\mathbf{g}) \in \mathbb{R}^{R \times R}$ are defined element-wise as
\begin{equation}
\begin{aligned}
s_r(\mathbf{g}) &= \left\|\mathbf{v}_r\right\| \prod_{n=1}^N \left\|g_{n,r}\right\|,\\
\mathbf{Z}_{r,s}(\mathbf{g}) &= A^{(0)}_{r,s}(\mathbf{g}) \, A^{(1)}_{r,s}(\mathbf{g}),\\
\mathbf{A}^{(0)}_{r,s}(\mathbf{g}) &= \cos(\theta_{\mathbf{v}_r, \mathbf{v}_s}) \prod_{n=1}^N \cos(\theta_{g_{n,r}, g_{n,s}}),\\
\mathbf{A}^{(1)}_{r,s}(\mathbf{g}) &= \sum_{n=1}^N \frac{\langle \nabla g_{n,r}, \nabla g_{n,s} \rangle}{\langle g_{n,r}, g_{n,s} \rangle},
\end{aligned}
\label{Theorem1b}
\end{equation}
where $\langle f, z \rangle := \int_0^1 f(x) z(x) \, dx$ denotes the standard $L^2$ inner product, and $\theta_{\cdot, \cdot}$ denotes the angle between two functions in the corresponding inner-product space.
\end{theorem}

The proof of the theorem above can be found in the Appendix. Notice that vector $\mathbf{s}\left(\mathbf{g}\right)\in\mathbb{R}_+^R$ holds the $L_2$ norms of each one of the $R$ tensor products/components of model (\ref{CPD}). According to (\ref{Theorem1b}), each element of $\mathbf{s}\left(\mathbf{g}\right)$ is defined as a product of the norms of the univariate functions $g_{n,r}$ and vector $\mathbf{v}_r$. Consequently, when most of the univariate functions have norms smaller than one, the resulting elements of $\mathbf{v}\left(\mathbf{g}\right)$ will be exponentially small. As a result, the Dirichlet energy can be exponentially small, unless the eigenvalues of the matrix $\mathbf{Z}\left(\mathbf{g}\right)$ are exponentially large. Among the components that constitute $\text{DE}(\mathbf{g})$, only the entries of $\mathbf{A}^{(0)}(\mathbf{g})$ involve products across all dimensions, yet these entries are bounded in absolute value by one.

Therefore, we argue that using the model in (\ref{CPD}) for learning tasks with Dirichlet energy-based regularization tends to promote functions with exponentially small values in $\mathbf{s}\left(\mathbf{g}\right)$. This is a pattern we have consistently observed in our experiments. In turn, this behavior induces a type of concentration phenomenon, where the energy becomes smoothly concentrated around the training points. Trying to eliminate the dependence on the vector $\mathbf{s}(\mathbf{g})$ shifts attention to the spectral properties of matrix $\mathbf{Z}\left(\mathbf{g}\right)$. In practice, we have observed that over many datasets and models such matrices are consistently positive semidefinite, i.e., all their eigenvalues are nonnegative. This implies that penalizing the sum of all positive eigenvalues as regularization corresponds to penalizing its trace as regularization. The downside of this approach is that each separable component of the model is penalized in isolation from all the rest, without regard to how the different components interact with each other. Next, we provide a more effective regularization approach that relies on local Dirichlet energies around the training points.

\subsection{Proposed Regularization}
\label{RegGen}

Another benefit of working with the model in (\ref{CPD}) is that it allow us to interpolate between sample-based approximations of Dirichlet Energy and its computation over the whole hypercube. This is accomplished by introducing a collection of small hypercubes centered at the training points, whose ``radius'' (half-side length) is a  tunable parameter. By adjusting this radius and adding the resulting local integrals, one can transition from purely pointwise (sample-based) approximations to increasingly global estimates that capture the geometry of the full domain. 

The resulting proposed regularization has the form
\begin{equation}
\text{LDE}_{\rho}\left(\mathbf{g}\right) := \sum_{m=1}^M\int_{\mathcal{B}_{\rho}\left(\mathbf{x}_m\right)}\left\|\nabla \mathbf{g}\left(\mathbf{x}\right)\right\|^2_Fd\mathbf{x},
\end{equation}
where $\mathcal{B}_{\rho}\left(\mathbf{x}_m\right)=\left\{\mathbf{x}\in\left[0,1\right]^N:\left\|\mathbf{x}_m-\mathbf{x}\right\|_{\infty}\leq \rho\right\}$. The advantage of this regularization is that it acts on the whole model and penalizes the localized Dirichlet energy at a union of balls centered at the training samples. This union can be considered as a rough approximation of the support of the density function from which the points are drawn from. This approximation has exponentially smaller volume than $\left[0,1\right]^{N}$. Hence, the promise of this approach stems from the fact that the Dirichlet energy is minimized over an exponentially smaller volume for all $\rho < 0.5$, strategically focusing on the support of the data-generating distribution. In the sequel, we consider $\text{LDE}_{\rho}\left(\mathbf{g}\right)$ as regularization for the tensor product of B-splines model.

Assuming a collection of data points $\mathcal{C} := \{(\mathbf{x}_m, y_m)\}_{m=1}^M$ and a loss function $\ell(y_m, \mathbf{g}(\mathbf{x}_m))$, we consider the regularized Empirical Risk Minimization problem
\begin{equation}
\begin{split}
\min_{\mathbf{g}} ~~& 
\frac{1}{M}\sum_{m=1}^M \ell\!\left(y_m, \mathbf{g}(\mathbf{x}_m)\right) 
\;+\; \lambda\, \mathrm{LDE}_{\rho}(\mathbf{g}) \\
\text{s.t.} ~~& ~~~~~~~~~~~~~~~~~~
\mathbf{g} \text{ admits } (\ref{CPD}),
\end{split}
\label{P1}
\end{equation}
where $\lambda > 0$ is a hyperparameter controlling the strength of the regularization.

In this work, we fit low-rank tensor product models within this framework and tune $\lambda$ using a penalty-method schedule. Specifically, we initialize $\lambda_0$ to a small value and update it according to $\lambda_{t+1} = h\,\lambda_t$ with $h>1$ each time the training algorithm converges. At the end of the procedure, we retain two candidate models: (i) the one achieving the best validation error overall, and (ii) the one achieving the best validation error after overfitting.

\subsection{Handling Incomplete Observations}

Missing entries in ${\bf x}$ pose significant challenges for most learning algorithms, particularly when the missingness is non-systematic or irregular. Simple mitigation strategies typically rely on imputation methods to fill in the missing values before applying standard learning algorithms. An easy way to do this is by using the averages of the missing features. However, the model proposed in (\ref{CPD}) can naturally accommodate missing data through its structural properties. Specifically, it allows for seamless integration over any subset of variables, enabling principled marginalization of unobserved dimensions without requiring explicit imputation. Without loss of generality, suppose that the first $D$ elements of a sample $\mathbf{x}\in\left[0,1\right]^N$ have not been observed and only the remaining entries $\mathbf{x}_{\text{obs}}$ have been observed. Then, the model in (\ref{CPD}) supports the following estimate 
\begin{equation}
\begin{split}
   \mathbf{y}_{\text{uni}} &=\int_{\left[0,1\right]^D}\mathbf{g}\left(x_1,\ldots,x_D, \mathbf{x}_{\text{obs}}\right)d\mathbf{x}\\
&=\sum_{r=1}^R\mathbf{v}_r\hspace{-1mm}\prod_{n=D+1}^N\hspace{-1mm}g_{n,r}\left(x_n\right)\prod_{d=1}^D\int_{\left[0,1\right]} g_{d,r}\left(x_d\right)dx_d.
   \end{split}
   \raisetag{20mm}
\end{equation}

A more sophisticated approach is to consider a simple model of a multivariate density function in the form of low rank tensor products \cite{karakasis2024multivariate}, 
\begin{equation}
    p\left(\mathbf{x}\right) = \sum_{s=1}^S w_s \prod_{n=1}^N p_{n,s}\left(x_n\right), 
\end{equation}
where $p_{n,s}\left(x_n\right):\left[0,1\right]\rightarrow \mathbb{R}_+$ denote bona fide univariate density functions and $\mathbf{w}\in\mathbb{R}_+^S$ denotes a probability mass function (PMF) of $S$ possible outcomes of a latent categorical variable. When $S$ is set to one, it is assumed that the observed features are independent, whereas $S>1$ means that those  features are \textit{conditionally independent} given the latent categorical variable. 

In this case, the marginalization can be performed using a density function learned from the dataset. Specifically, the resulting estimator  is given by
\begin{equation}
\begin{split}
 &~\mathbf{y}_{\text{pdf}} 
 = \mathbb{E}_p\left[\mathbf{g}~\big|~\text{observed}=\mathbf{x}_{\text{obs}}\right]\\
 &=\int_{\left[0,1\right]^D}\hspace{-6mm}\mathbf{g}\left(x_1,\ldots,x_D,\mathbf{x}_{\text{obs}}\right)p\left(x_1,\ldots,x_D\big| \mathbf{x}_{\text{obs}}\right)dx_1\dots dx_D\\  
&=\int_{\left[0,1\right]^D}\hspace{-6mm}\mathbf{g}\left(x_1,\ldots,x_D,\mathbf{x}_{\text{obs}}\right)\frac{p\left(x_1,\ldots,x_D,\mathbf{x}_{\text{obs}}\right)}{p\left(\mathbf{x}_{\text{obs}}\right)}dx_1\dots dx_D\\  &=\frac{\sum_{r=1}^R\sum_{s=1}^S\mathbf{v}_r~w_s\prod_{n=D+1}^Ng_{n,r}\left(x_n\right)\prod_{d=1}^D\left<g_{d,r}p_{d,s}\right>_{\left[0,1\right]}}{\sum_{t=1}^Sw_t\prod_{v=D+1}^Np_{v,t}\left(x_v\right)}.
 \raisetag{36mm}
\end{split}
\end{equation}
Note that in the ideal case where the model for $p$ matches the true data-generating distribution, the estimator in (12) coincides with the optimal mean squared error (MMSE) estimator.

\section{Experimental Results}

In this section, we evaluate the performance of the proposed framework using B-splines on supervised learning tasks, considering both fully observed datasets and scenarios with incomplete observations at inference time. 

We use six standard datasets spanning classification and regression tasks. For classification, we consider the Ionosphere (Ion)\footnote{\url{https://archive.ics.uci.edu/ml/datasets/ionosphere}} dataset, containing 34 features describing radar returns for detecting structures in the ionosphere, and the Breast Cancer Wisconsin (BCW)\footnote{\url{https://archive.ics.uci.edu/ml/datasets/Breast+Cancer+Wisconsin+(Original)}} dataset with 30 features for classifying malignant and benign tumors. For regression, we use the Diabetes\footnote{\url{https://scikit-learn.org/stable/modules/generated/sklearn.datasets.load_diabetes.html}} dataset from \texttt{scikit-learn}, which has 10 features predicting disease progression, the Yacht Hydrodynamics\footnote{\url{https://archive.ics.uci.edu/ml/datasets/yacht+hydrodynamics}} dataset with 6 features modeling yacht drag coefficients, the Physico\footnote{\url{https://archive.ics.uci.edu/ml/datasets/Physicochemical+Properties+of+Protein+Tertiary+Structure}} dataset with 9 features related to protein structure, and the Sarcos\footnote{\url{https://www.gaussianprocess.org/gpml/data/}} robot arm dataset with 21 features for inverse dynamics prediction. For each dataset, we generate three random splits into training, validation, and test sets. Table~1 summarizes their key characteristics.

As baselines, we consider neural networks with one or two hidden layers, regularized using the average Frobenius norm of the Jacobian computed over the training samples. For single-hidden-layer networks, we use widths of 1000, 5000, and 20000 neurons, while for two-layer networks, we choose equal widths to match the total number of parameters of the corresponding single-layer models. For the tensor–product of B-splines (TPBS) models, we fix the number of knots per dimension to 100 and vary the rank so that the total number of parameters is comparable to the neural network counterparts. Additionally, to assess the effectiveness of these regularization approaches, we present results for both models with and without their use. It should be noted that the case of TPBS models without regularization reduces to the approach proposed by \cite{govindarajan2022regression}.
Finally, we use AdamW \cite{loshchilov2017decoupled} with weight decay equal to 0, eps equal to 1e-4, and manually tuned learning rates for training all models.

For comparison, we retain the models achieving the best validation performance (i) in general, and (ii) after controlled overfitting while gradually increasing the regularization hyperparameters. Here, overfitting refers to the case when the training error falls below a prescribed threshold. This analysis allows us to explore whether controlled overfitting can be beneficial and how different model classes respond to overfitting and the use of regularization.  

\begin{table}[ht!]
\centering
\begin{tabular}{lccccc}
\hline
\textbf{Dataset} & \textbf{Dim} & \textbf{Task} & \textbf{Train} & \textbf{Val} & \textbf{Test} \\
\hline
Ion      & 34 & Classification & 160 & 100 & 91 \\
BCW      & 30 & Classification & 260 & 100 & 200 \\
Diabetes & 10 & Regression     & 200 & 100 & 142 \\
Yacht    & 6  & Regression     & 150 & 58 & 100 \\
Physico  & 9  & Regression     & 250 & 500 & 1000 \\
Sarcos   & 21 & Regression     & 250 & 500 & 4449 \\
\hline
\end{tabular}
\caption{Dataset statistics: dimensionality, task type, and number of samples in training/validation/testing splits.}
\end{table}

 \begin{table*}[t!]
\centering
\resizebox{0.8\textwidth}{!}{%
\begin{tabular}{lcccc}
\toprule
Dataset & NN (best val) & NN (overfit) & TPBS (best val) & TPBS (overfit) \\
\midrule
Ion (34D - $\uparrow$) 
& 0.949 $\pm$ 0.026 (NR)
& \textbf{0.952 $\pm$ 0.014} (R)
& 0.930 $\pm$ 0.029 (R)
& 0.938 $\pm$ 0.010 (R)\\
\midrule
BCW (30D - $\uparrow$) 
& \textbf{0.975 $\pm$ 0.007} (NR)
& 0.958 $\pm$ 0.012 (NR)
& 0.965 $\pm$ 0.014 (R)
& 0.965 $\pm$ 0.014 (R)\\
\midrule
Diabetes (10D - $\downarrow$) 
& \textbf{0.101 $\pm$ 0.010} (R)
& 0.223 $\pm$ 0.037 (R)
& 0.134 $\pm$ 0.005 (R)
& 0.173 $\pm$ 0.019 (R)\\
\midrule
Yacht (6D - $\downarrow$) 
& \textbf{0.001 $\pm$ 0.000} (R)
& \textbf{0.001 $\pm$ 0.000} (R)
& \textbf{0.001 $\pm$ 0.000} (R)
& \textbf{0.001 $\pm$ 0.000} (R)\\
\midrule
Physico (9D - $\downarrow$)
&  \textbf{0.263 $\pm$ 0.010} (NR)
& 0.533 $\pm$ 0.075 (NR)
& 0.340 $\pm$ 0.012 (R)
& 0.336 $\pm$ 0.013 (R)\\
\midrule
Sarcos (21D - $\downarrow$)
& \textbf{0.049 $\pm$ 0.001} (NR)
& 0.060 $\pm$ 0.005 (NR)
& 0.240 $\pm$ 0.042 (R)
& 0.243 $\pm$ 0.046 (R) \\
\bottomrule
\end{tabular}%
}
\caption{Summary of best model test performance across all datasets. NN values refer to the best-performing neural network (across all combinations of hidden-layer counts and whether regularization was applied) before and after overfitting. TPBS values refer to the best low-rank tensor product B-spline model across ranks and regularization options. (R) and (NR) denote whether the best performance was achieved by a model with or without regularization, respectively. For datasets with an arrow pointing upwards, the task under consideration is classification, and the reported metric is classification accuracy. Hence, the higher, the better. For datasets with an arrow pointing downwards, the task under consideration is regression, and the reported metric is relative mean square error. Hence, the lower, the better.}\vspace{-3mm}
 \label{tab:FinalA}
\end{table*}

\subsection{Inference with full datasets}

In this subsection, we present the results that emerged after training TPBS and neural network models and using fully observed samples for inference. In the appendix, Tables (\ref{tab:Ion})-(\ref{tab:sarcos}) present detailed results for all datasets and model configurations, and in Table (\ref{tab:FinalA}), we summarize the best-performing models across all experiments for a concise comparison. In Table (\ref{tab:FinalA}), we observe that neural networks and TPBS models achieve comparable performance on the first four datasets, with neural networks leading in three of them by a small margin. Overfitting consistently harms neural network performance in all but two datasets. In contrast, TPBS models exhibit notably more robust behavior: in half of the datasets, overfitting does not degrade performance at all—indeed, in three cases it even improves it. In the remaining datasets, the negative impact of overfitting is far milder compared to neural networks. For example, in the Diabetes and Physico datasets, overfitting nearly doubles the relative mean squared error for neural networks, whereas TPBS performance remains largely unchanged. We also observe that the use of regularization consistently improves TPBS performance across all datasets.

When comparing the two methods specifically in the overfitting regime, TPBS models outperform neural networks in four out of six datasets, with only one of the remaining datasets showing a substantial advantage for neural networks. Therefore, in applications where perfect memorization of the training data is desired—such as associative memory models, error-correcting code reconstruction, or dictionary-based compression—TPBS models offer a compelling alternative.

\subsection{Inference with incomplete datasets}

Tables (\ref{tab:ioninc})–(\ref{tab:yachtinc}) present the results obtained in the inference from incomplete observations setting, evaluated under both overfitted and non-overfitted models. 
For each dataset, we consider scenarios in which 2, 3, or 4 entries of every observation are randomly hidden. In all cases, we reuse the models selected in the previous subsection—those trained on fully observed samples together with their associated regularization schemes; the proposed LDE $\rho$ regularization for TPBS models, and the averaged Jacobian norm for neural networks.

For the neural network models, we impute the missing entries using the empirical mean of each feature and then apply the pretrained networks directly. For the tensor product of B-splines (TPBS) models, we evaluate four different approaches.
The first approach mirrors the neural network case: we impute missing entries with the corresponding feature means and use the TPBS model trained on fully observed data. The second approach replaces each missing entry with the integral of the corresponding learned univariate function, following the estimator described in relation (10). The third and fourth approaches substitute each missing entry with the inner product between the learned low-rank density model and the corresponding univariate function, using ranks $S=1$ and $S=5$, respectively. These two methods correspond to estimators of the form given in relation (12). In all cases, the resulting weighted integrals or inner products serve as surrogates for the missing values.

For the Ion dataset (Table \ref{tab:ioninc}), we observe that neural networks and TPBS models perform comparably in both the overfitting and non-overfitting regimes across all scenarios of missing entries. Overfitting seems to help both models in some cases. The performance gap grows in the overfitting case where the best performance across the entire table (for 2, 3, and 4 missing elements) is achieved by our marginalized method in the case of 4 missing elements. This result may appear counterintuitive, since a larger number of missing entries would typically be expected to degrade performance. A plausible explanation
is the presence of noise in the data: in this regime, relying on incomplete observations may be more robust, as the model depends less on potentially noisy measurements and more
on the underlying data statistics.

For the BCW dataset (Table \ref{tab:bcwinc}), we observe that TPBS models outperform neural networks in all but the first scenario of two random missing entries in the non-overfitting regime. Moreover, TBPS appears to be noticeably insensitive to overfitting, while in neural networks the difference is still small but more apparent. The performance gap between the two models appears in the overfitting case and is constant across all scenarios of missing entries. Again, regularization consistently benefits the TPBS models in all cases and scenarios, but improves the neural network performance only in the overfitting case.

For the Sarcos dataset (Table \ref{tab:sarcosinc}), we observe that neural networks outperform TPBS models in both the overfitting and non-overfitting regimes across all scenarios of missing entries. The performance gap narrows in the overfitting case and in the scenario with 4 missing entries. While regularization does not always improve neural network performance, it consistently benefits TPBS models. For the Physico dataset (Table \ref{tab:physicoinc}), TPBS models outperform neural networks in both regimes and across all missing-entry scenarios. Interestingly, overfitting consistently improves TPBS performance, whereas it severely degrades neural network performance. Again, regularization is consistently beneficial for TPBS models, while its effect on neural networks is actually detrimental (best performing model does not use regularization in all cases considered). 

For the Diabetes dataset (Table \ref{tab:diabetesinc}), neural networks slightly outperform TPBS models in the non-overfitting regime, while TPBS models clearly lead in the overfitting regime. Overfitting has a mild effect on TPBS models but significantly harms neural networks. Here, regularization consistently improves neural network performance, and for TPBS models, it helps in the overfitting regime but has little effect in the non-overfitting case. Finally, for the Yacht dataset (Table \ref{tab:yachtinc}), TPBS models are the clear winners in both regimes and across all missing-entry scenarios. Overfitting has minimal impact on either method. Regularization consistently benefits neural networks with two hidden layers and TPBS models across all regimes and scenarios.

Overall, these results indicate that LDE-regularized TPBS models exhibit greater robustness to overfitting and consistently benefit from regularization, whereas neural networks are more sensitive to overfitting and show less predictable gains from regularization.

\begin{table*}[ht]
\centering
\scalebox{0.88}{
\begin{tabular}{lccc}
\toprule
\textbf{Method } & 2 missing entries & 3 missing entries & 4 missing entries \\
\midrule
\multicolumn{4}{l}{\textbf{Best based on Validation Error}} \\
\midrule
NN 1HL  & \textbf{0.941 $\pm$ 0.014} (reg, N5000) & \textbf{0.941} $\pm$ \textbf{0.019} (no reg, N1000) & \textbf{0.941 $\pm$ 0.014} (no reg, N20000) \\
NN 2HL  & 0.941 $\pm$ 0.029 (no reg, N1000)& 0.938 $\pm$ 0.010 (no reg, N20000) & 0.938 $\pm$ 0.023 (reg, N1000) \\
\hline
TPBS  & $0.927 \pm 0.014$ (LDE $\rho$=0.01, R11) & $0.930 \pm 0.029$ (LDE $\rho$=0.01, R227) & $0.934 \pm 0.024$ (LDE $\rho$=0.1, R227) \\
TPBS MARG  & \textbf{0.941 $\pm$ 0.029} (LDE $\rho$=0.1, R227) & \textbf{0.938 $\pm$ 0.034} (LDE $\rho$=0.1, R227) & \textbf{0.941 $\pm$ 0.021} (LDE $\rho$=0.1, R227) \\
TPBS PDF S1  & $0.934 \pm 0.039$ (LDE $\rho$=0.1, R227) & $0.930 \pm 0.021$ (LDE $\rho$=0.01, R11) & $0.934 \pm 0.024$ (LDE $\rho$=0.1, R227) \\
TPBS PDF S5  & $0.938 \pm 0.034$ (LDE $\rho$=0.1, R227) & $0.934 \pm 0.031$ (LDE $\rho$=0.01, R227) & $0.934 \pm 0.024$ (LDE $\rho$=0.1, R227) \\
\midrule
\multicolumn{4}{l}{\textbf{After Overfitting}} \\
\midrule
NN 1HL  & \textbf{0.952 $\pm$ 0.027} (reg, N20000) & \textbf{0.941 $\pm$ 0.034} (reg, N20000) & \textbf{0.938 $\pm$ 0.036} (reg, N20000)\\
NN 2HL  & $0.949 \pm 0.029$ (reg, N1000) & \textbf{0.941 $\pm$ 0.034} (reg, N1000) & $0.930 \pm 0.034$ (reg, N1000) \\
\midrule
TPBS & $0.938 \pm 0.010$ (LDE $\rho$=0.01, R57) & $0.938 \pm 0.019$ (LDE $\rho$=0.01, R227) & $0.934 \pm 0.018$ (LDE $\rho$=0.01, R57) \\
TPBS MARG & \textbf{0.945 $\pm$ 0.024} (LDE $\rho$=0.1, R227) & \textbf{0.941 $\pm$ 0.005} (LDE $\rho$=0.2, R227) & \textbf{0.956 $\pm$ 0.018} (LDE $\rho$=0.01, R57) \\
TPBS PDF S1 & $0.938 \pm 0.034$ (LDE $\rho$=0.1, R227) & $0.934 \pm 0.018$ (LDE $\rho$=0.01, R227) & $0.938 \pm 0.021$ (LDE $\rho$=0.01, R57) \\
TPBS PDF S5  & $0.934 \pm 0.009$ (LDE $\rho$=0.2, R227) & $0.934 \pm 0.009$ (LDE $\rho$=0.2, R227) & $0.938 \pm 0.021$ (LDE $\rho$=0.01, R57) \\
\bottomrule
\end{tabular}}
\caption{Ion dataset results. Each row corresponds to a method and each column shows the number of random missing entries. The first block of rows corresponds to the models that achieved the best validation error during training with fully observed samples. The second block to models that achieved the best validation error during training with fully observed samples, but in the overfitting regime. The task here is classification, and the reported metric is classification accuracy, so the higher the better. The values in the parentheses indicate what was the rank of the best model or the number of neurons, respectively, as well as the regularization setup that gave that model. LDE $\rho$=x stands for using the proposed regularization with radius x.}\vspace{-2mm}
\label{tab:ioninc}
\end{table*}

\begin{table*}[h!]
\centering
\scalebox{0.88}{
\begin{tabular}{lccc}
\toprule
\textbf{Method } & 2 missing entries & 3 missing entries & 4 missing entries \\
\midrule
\multicolumn{4}{l}{\textbf{Best based on Validation Error}} \\
\midrule
NN 1HL  & 0.965 $\pm$ 0.024 (no reg, N20000) & $0.960 \pm 0.018$ (no reg, N1000) & $0.952 \pm 0.016$ (reg, N20000) \\
NN 2HL  & \textbf{0.973 $\pm$ 0.008} (no reg, N20000)& \textbf{0.963 $\pm$ 0.009} (no reg, N1000) & \textbf{0.965 $\pm$ 0.007} (no reg, N5000) \\
\hline
TPBS  & $0.967 \pm 0.012$ (LDE $\rho$=0.01, R229) & $0.962 \pm 0.023$ (LDE $\rho$=0.01, R229) & $0.962 \pm 0.016$ (LDE $\rho$=0.01, R229) \\
TPBS MARG  & \textbf{0.968 $\pm$ 0.013} (LDE $\rho$=0.1, R229) & \textbf{0.967 $\pm$ 0.022} (LDE $\rho$=0.01, R229) & $0.963 \pm 0.014$ (LDE $\rho$=0.01, R229) \\
TPBS PDF S1  & \textbf{0.968 $\pm$ 0.013} (LDE $\rho$=0.1, R229) & $0.962 \pm 0.012$ (LDE $\rho$=0.1, R229) & \textbf{0.967 $\pm$ 0.015} (LDE $\rho$=0.1, R229) \\
TPBS PDF S5  & $0.967 \pm 0.019$ (LDE $\rho$=0.01, R229) & $0.963 \pm 0.021$ (LDE $\rho$=0.1, R229) & $0.955 \pm 0.032$ (LDE $\rho$=0.1, R229) \\
\midrule
\multicolumn{4}{l}{\textbf{After Overfitting}} \\
\midrule
NN 1HL  & 0.947 $\pm$ 0.015 (reg, N5000) & 0.948 $\pm$ 0.027 (reg, N1000) & \textbf{0.953 $\pm$ 0.013} (reg, N1000)\\
NN 2HL  & \textbf{0.957 $\pm$ 0.019} (reg, N20000) & \textbf{0.955 $\pm$ 0.018} (reg, N5000) & $0.950 \pm 0.016$ (reg, N20000) \\
\midrule
TPBS & $0.967 \pm 0.012$ (LDE $\rho$=0.01, R229) & $0.963 \pm 0.021$ (LDE $\rho$=0.01, R229) & $0.960 \pm 0.019$ (LDE $\rho$=0.01, R229) \\
TPBS MARG & \textbf{0.968 $\pm$ 0.013} (LDE $\rho$=0.1, R229) & \textbf{0.968 $\pm$ 0.020} (LDE $\rho$=0.01, R229) & \textbf{0.963 $\pm$ 0.014} (LDE $\rho$=0.01, R229) \\
TPBS PDF S1 & \textbf{0.968 $\pm$ 0.013} (LDE $\rho$=0.1, R229) & 0.963 $\pm$ 0.018 (LDE $\rho$=0.01, R229) & $0.962 \pm 0.016$ (LDE $\rho$=0.01, R229) \\
TPBS PDF S5  & $0.967 \pm 0.019$ (LDE $\rho$=0.01, R229) & $0.963 \pm 0.020$ (LDE $\rho$=0.1, R229) & $0.957 \pm 0.030$ (LDE $\rho$=0.1, R229) \\
\bottomrule
\end{tabular}}
\caption{BCW dataset results. Each row corresponds to a method and each column shows the number of random missing entries. The first block of rows corresponds to the models that achieved the best validation error during training with fully observed samples. The second block to models that achieved the best validation error during training with fully observed samples, but in the overfitting regime. The task here is classification, and the reported metric is classification accuracy, so the higher the better. The values in the parentheses indicate what was the rank of the best model or the number of neurons, respectively, as well as the regularization setup that gave that model. LDE $\rho$=x stands for using the proposed regularization with radius x.}\vspace{-2mm}
\label{tab:bcwinc}
\end{table*}

\begin{table*}[ht]
\centering
\scalebox{0.88}{
\begin{tabular}{lccc}
\toprule
\textbf{Method } & 2 missing entries & 3 missing entries & 4 missing entries \\
\midrule
\multicolumn{4}{l}{\textbf{Best based on Validation Error}} \\
\midrule
NN 1HL  & \textbf{0.147 $\pm$ 0.016} (no reg, N20000) & $0.199 \pm 0.008$ (no reg, N20000) & $0.249 \pm 0.008$ (no reg, N20000) \\
NN 2HL  & $0.147 \pm 0.019$ (reg, N20000)& \textbf{0.192 $\pm$ 0.004} (reg, N20000) & \textbf{0.238 $\pm$ 0.012} (no reg, N20000) \\
\hline
TPBS  & $0.280 \pm 0.038$ (LDE $\rho$=0.01, R57) & $0.298 \pm 0.040$ (LDE $\rho$=0.01, R57) & $0.323 \pm 0.035$ (LDE $\rho$=0.01, R57) \\
TPBS MARG  & $0.298 \pm 0.029$ (LDE $\rho$=0.01, R57) & $0.325 \pm 0.028$ (LDE $\rho$=0.01, R57) & $0.360 \pm 0.020$ (LDE $\rho$=0.01, R57) \\
TPBS PDF S1  & $0.277 \pm 0.037$ (LDE $\rho$=0.01, R57) & $0.292 \pm 0.036$ (LDE $\rho$=0.01, R57) & $0.316 \pm 0.031$ (LDE $\rho$=0.01, R57) \\
TPBS PDF S5  & $0.275 \pm 0.038$ (LDE $\rho$=0.01, R57) & $0.292 \pm 0.037$ (LDE $\rho$=0.01, R57) & $0.312 \pm 0.031$ (LDE $\rho$=0.01, R57) \\
\midrule
\multicolumn{4}{l}{\textbf{After Overfitting}} \\
\midrule
NN 1HL  & \textbf{0.171 $\pm$ 0.005} (no reg, N20000) & \textbf{0.233 $\pm$ 0.025} (wo reg, N20000) & \textbf{0.288 $\pm$ 0.023} (no reg, N20000)\\
NN 2HL  & $0.184 \pm 0.010$ (no reg, N20000) & $0.238 \pm 0.016$ (no reg, N20000) & $0.289 \pm 0.013$ (no reg, N20000) \\
\midrule
TPBS & $0.283 \pm 0.042$ (LDE $\rho$=0.01, R57) & $0.300 \pm 0.043$ (LDE $\rho$=0.01, R57) & $0.325 \pm 0.038$ (LDE $\rho$=0.01, R57) \\
TPBS MARG & $0.300 \pm 0.032$ (LDE $\rho$=0.01, R57) & $0.326 \pm 0.030$ (LDE $\rho$=0.01, R57) & $0.362 \pm 0.022$ (LDE $\rho$=0.01, R57) \\
TPBS PDF S1 & $0.279 \pm 0.040$ (LDE $\rho$=0.01, R57) & $0.294 \pm 0.039$ (LDE $\rho$=0.01, R57) & $0.318 \pm 0.034$ (LDE $\rho$=0.01, R57) \\
TPBS PDF S5  & $0.277 \pm 0.041$ (LDE $\rho$=0.01, R57) & $0.294 \pm 0.040$ (LDE $\rho$=0.01, R57) & $0.314 \pm 0.034$ (LDE $\rho$=0.01, R57) \\
\bottomrule
\end{tabular}}
\caption{Sarcos dataset results. Each row corresponds to a method and each column shows the number of random missing entries. The first block of rows corresponds to the models that achieved the best validation error during training with fully observed samples. The second block to models that achieved the best validation error during training with fully observed samples, but in the overfitting regime. The task here is regression, and the reported metric is relative mean squared error, so the lower the better. The values in the parentheses indicate what was the rank of the best model or the number of neurons, respectively, as well as the regularization setup that gave that model. LDE $\rho$=x stands for using the proposed regularization with radius x.}
\label{tab:sarcosinc}
\end{table*}

\begin{table*}[h!]
\centering
\scalebox{0.88}{
\begin{tabular}{lccc}
\toprule
\textbf{Method} & 2 missing entries & 3 missing entries & 4 missing entries\\
\midrule
\midrule
\multicolumn{4}{l}{\textbf{Best based on Validation Error}} \\
\midrule
NN 1HL  & 0.407 $\pm$ 0.012 (no reg, N20000) & $0.455 \pm 0.014$ (no reg, N20000) & $0.475 \pm 0.006$ (no reg, N20000) \\
NN 2HL  & $0.416 \pm 0.037$ (no reg, N20000)& 0.463 $\pm$ 0.030 (reg, N20000) & 0.468 $\pm$ 0.036 (no reg, N5000) \\
\midrule
TPBS & 0.351 $\pm$ 0.005 (LDE $\rho$=0.2, R64) & $0.360 \pm 0.008$ (LDE $\rho$=0.2, R64) & $0.364 \pm 0.004$ (LDE $\rho$=0.2, R64) \\
TPBS MARG & $0.357 \pm 0.000$ (LDE $\rho$=0.2, R13) & $0.359 \pm 0.003$ (LDE $\rho$=0.2, R13) & $0.363 \pm 0.003$ (LDE $\rho$=0.2, R13) \\
TPBS PDF S1  & $0.352 \pm 0.003$ (LDE $\rho$=0.1, R64) & $0.358 \pm 0.006$ (LDE $\rho$=0.1, R64) & $0.358 \pm 0.001$ (no reg, R64) \\
TPBS PDF S5 & \textbf{0.340 $\pm$ 0.019} (LDE $\rho$=0.01, R64) & \textbf{0.346 $\pm$ 0.015} (LDE $\rho$=0.01, R64) & \textbf{0.352 $\pm$ 0.016} (LDE $\rho$=0.01, R64) \\
\midrule
\multicolumn{4}{l}{\textbf{After Overfitting}} \\
\midrule
NN 1HL  & $0.995 \pm 0.012$ (no reg, N1000) & $0.995 \pm 0.012$ (no reg, N1000) & $0.995 \pm 0.012$ (no reg, N1000)\\
NN 2HL & $0.991 \pm 0.001$ (no reg, N1000) & $0.991 \pm 0.001$ (no reg, N1000) & $0.991 \pm 0.001$ (no reg, N1000) \\
\midrule
TPBS & $0.412 \pm 0.019$ (LDE $\rho$=0.1, R64) & $0.448 \pm 0.047$ (LDE $\rho$=0.1, R64) & $0.455 \pm 0.042$ (LDE $\rho$=0.1, R64) \\
TPBS MARG  & $0.402 \pm 0.063$ (LDE $\rho$=0.2, R64) & $0.408 \pm 0.046$ (LDE $\rho$=0.2, R64) & $0.419 \pm 0.043$ (LDE $\rho$=0.2, R64) \\
TPBS PDF S1  & $0.366 \pm 0.007$ (LDE $\rho$=0.1, R64) & $0.372 \pm 0.018$ (LDE $\rho$=0.1, R64) & 0.370 $\pm$ 0.027 (LDE $\rho$=0.1, R64) \\
TPBS PDF S5  & \textbf{0.330 $\pm$ 0.014} (LDE $\rho$=0.01, R64) & \textbf{0.334 $\pm$ 0.007} (LDE $\rho$=0.01, R64) & \textbf{0.347 $\pm$ 0.014} (LDE $\rho$=0.01, R64) \\
\bottomrule
\end{tabular}}
\caption{Physico dataset results. Each row corresponds to a method and each column shows the number of random missing entries. The first block of rows corresponds to the models that achieved the best validation error during training with fully observed samples. The second block to models that achieved the best validation error during training with fully observed samples, but in the overfitting regime. The task here is regression, and the reported metric is relative mean squared error, so the lower the better. The values in the parentheses indicate what was the rank of the best model or the number of neurons, respectively, as well as the regularization setup that gave that model. LDE $\rho$=x stands for using the proposed regularization with radius x.}
\label{tab:physicoinc}
\end{table*}

\begin{table*}[ht]
\centering
\scalebox{0.88}{
\begin{tabular}{lccc}
\toprule
\textbf{Method} & 2 missing entries & 3 missing entries & 4 missing entries\\
\midrule
\midrule
\multicolumn{4}{l}{\textbf{Best based on Validation Error}} \\
\midrule
NN 1HL  & \textbf{0.109 $\pm$ 0.008} (reg, N1000) & $0.116 \pm 0.005$ (reg, N1000) & \textbf{0.123 $\pm$ 0.006} (reg, N1000) \\
NN 2HL  & $0.109 \pm 0.008$ (reg, N1000)& \textbf{0.116 $\pm$ 0.004 }(reg, N5000) & 0.123 $\pm$ 0.008 (reg, N20000) \\
\hline
TPBS  & 0.148 $\pm$ 0.001 (no reg, R250) & $0.158 \pm 0.004$ (no reg, R250) & $0.157 \pm 0.007$ (no reg, R250) \\
TPBS MARG  & 0.143 $\pm$ 0.006 (no reg, R250) & $0.150 \pm 0.007$ (no reg, R250) & $0.148 \pm 0.010$ (no reg, R250) \\
TPBS PDF S1  & $0.143 \pm 0.006$ (no reg, R250) & $0.146 \pm 0.006$ (LDE $\rho$=0.1, R250) & $0.147 \pm 0.010$ (no reg, R250) \\
TPBS PDF S5  & $0.143 \pm 0.006$ (no reg, R250) & $0.146 \pm 0.006$ (LDE $\rho$=0.1, R250) & 0.147 $\pm$ 0.010 (no reg, R250) \\

\midrule
\multicolumn{4}{l}{\textbf{After Overfitting}} \\
\midrule
NN 1HL  & $0.252 \pm 0.029$ (reg, N20000) & $0.240 \pm 0.027$ (reg, N20000) & $0.242 \pm 0.019$ (reg, N20000)\\
NN 2HL  & $0.363 \pm 0.139$ (reg, N20000) & $0.338 \pm 0.083$ (reg, N20000) & $0.388 \pm 0.163$ (reg, N20000) \\
\midrule
TPBS  & $0.189 \pm 0.006$ (LDE $\rho$=0.01, R250) & $0.197 \pm 0.008$ (LDE $\rho$=0.01, R250) & $0.201 \pm 0.013$ (LDE $\rho$=0.01, R250) \\
TPBS MARG  & $0.170 \pm 0.002$ (LDE $\rho$=0.01, R250) & $0.184 \pm 0.013$ (LDE 0.01, R250) & $0.184 \pm 0.010$ (LDE 0.01, R250) \\
TPBS PDF S1  & \textbf{0.166 $\pm$ 0.002} (LDE $\rho$=0.01, R250) & \textbf{0.169 $\pm$ 0.014} (LDE $\rho$=0.01, R250) & \textbf{0.172 $\pm$ 0.008} (LDE $\rho$=0.01, R250) \\
TPBS PDF S5  & 0.167 $\pm$ 0.001 (LDE $\rho$=0.01, R250) & 0.170 $\pm$ 0.013 (LDE $\rho$=0.01, R250) & \textbf{0.172 $\pm$ 0.008} (LDE $\rho$=0.01, R250) \\
\bottomrule
\end{tabular}}
\caption{Diabetes dataset results. Each row corresponds to a method and each column shows the number of random missing entries. The first block of rows corresponds to the models that achieved the best validation error during training with fully observed samples. The second block to models that achieved the best validation error during training with fully observed samples, but in the overfitting regime. The task here is regression, and the reported metric is relative mean squared error, so the lower the better. The values in the parentheses indicate what was the rank of the best model or the number of neurons, respectively, as well as the regularization setup that gave that model. LDE $\rho$=x stands for using the proposed regularization with radius x.}
\label{tab:diabetesinc}
\end{table*}

\begin{table*}[h!]
\centering
\scalebox{0.88}{
\begin{tabular}{lccc}
\toprule
\textbf{Method} & 2 missing entries& 3 missing entries & 4 missing entries \\
\midrule
\midrule
\multicolumn{4}{l}{\textbf{Best based on Validation Error}} \\
\midrule
NN 1HL  & $0.141 \pm 0.101$ (no reg, N20000) & $0.542 \pm 0.057$ (no reg, N20000) & $0.575 \pm 0.039$ (no reg, N20000)\\
NN 2HL  & $0.140 \pm 0.102$ (reg, N20000) & $0.541\pm 0.065$ (reg, N1000) & $0.572 \pm 0.044$ (reg, N1000) \\
\hline
TPBS  & $0.127 \pm 0.101$ (LDE $\rho$=0.1, R277) & $0.505 \pm 0.055$ (LDE $\rho$=0.1, R277) & $0.528 \pm 0.029$ (LDE $\rho$=0.1, R277) \\
TPBS MARG  & 0.144 $\pm$ 0.067 (LDE $\rho$=0.2, R277) & $0.442 \pm 0.064$ (LDE $\rho$=0.2, R277) & $0.522 \pm 0.048$ (LDE $\rho$=0.2, R277) \\
TPBS PDF S1  & \textbf{0.114 $\pm$ 0.070} (LDE $\rho$=0.2, R69) & \textbf{0.399 $\pm$ 0.071} (no reg, R14) & \textbf{0.487 $\pm$ 0.056} (LDE $\rho$=0.01, R14) \\
TPBS PDF S5  & $0.118 \pm 0.077$ (LDE $\rho$=0.2, R277) & $0.404 \pm 0.081$ (no reg, R14) & \textbf{0.497 $\pm$ 0.054} (LDE $\rho$=0.01, R14) \\

\midrule
\multicolumn{4}{l}{\textbf{After Overfitting}} \\
\midrule
NN 1HL  & $0.141 \pm 0.101$ (no reg, N20000) & $0.542 \pm 0.057$ (no reg, N20000) & $0.575 \pm 0.039$ (no reg, N20000)\\
NN 2HL  & $0.140 \pm 0.101$ (reg, N20000) & $0.540\pm 0.064$ (reg, N1000) & $0.571 \pm 0.043$ (reg, N1000) \\
\midrule
TPBS  & $0.127 \pm 0.101$ (LDE $\rho$=0.1, R277) & $0.505 \pm 0.055$ (LDE $\rho$=0.1, R277) & $0.528 \pm 0.029$ (LDE $\rho$=0.1, R277) \\
TPBS MARG  & $0.144 \pm 0.067$ (LDE $\rho$=0.2, R277) & $0.442 \pm 0.063$ (LDE $\rho$=0.2, R277) & $0.523 \pm 0.046$ (LDE $\rho$=0.2, R277) \\
TPBS PDF S1  & \textbf{0.114 $\pm$ 0.070} (LDE $\rho$=0.2, R69) &0.409 $\pm$ 0.066 (LDE $\rho$=0.2, R14) & 0.487 $\pm$ 0.055 (LDE $\rho$=0.01, R14) \\
TPBS PDF S5  & 0.118 $\pm$ 0.077 (LDE $\rho$=0.2, R277) & \textbf{0.405 $\pm$ 0.080} (no reg, R14) & \textbf{0.46 $\pm$ 0.049} (LDE $\rho$=0.2, R14) \\
\bottomrule
\end{tabular}}
\caption{Yacht dataset results. Each row corresponds to a method and each column shows the number of random missing entries. The first block of rows corresponds to the models that achieved the best validation error during training with fully observed samples. The second block to models that achieved the best validation error during training with fully observed samples, but in the overfitting regime. The task here is regression, and the reported metric is relative mean squared error, so the lower the better. The values in the parentheses indicate what was the rank of the best model or the number of neurons, respectively, as well as the regularization setup that gave that model. LDE $\rho$=x stands for using the proposed regularization with radius x.}
\label{tab:yachtinc}
\end{table*}

\section{Conclusion}

In this work, we analyzed low-rank tensor-product of B-splines models through the lens of Dirichlet energy. We showed that the Dirichlet energy of these models can be computed in closed form, providing insight into their internal mechanisms. While global Dirichlet energy can be exponentially small even under perfect interpolation, making it unsuitable as a regularization, our proposed approach, relying on local Dirichlet energies, effectively controls smoothness near the training points, capturing the support of the data distribution. We developed two estimators that leverage pretrained TPBS models to perform accurate inference from incomplete data, naturally combining regularization and low-rank structure. Comparative experiments with neural networks revealed that TPBS models consistently outperform neural networks in the overfitting regime and in scenarios with missing entries, while achieving comparable performance in most other cases. These results demonstrate the robustness of TPBS models to overfitting, the consistent benefits of LDE-regularization, and their strong potential for inference under missing data scenarios. Moreover, TPBS models seem to be particularly suitable for applications requiring perfect memorization of training data, while also accommodating incomplete observations, offering a flexible alternative to neural networks.

\bibliography{main_paper}
\bibliographystyle{icml2026}

\newpage
\appendix
\onecolumn

\section{Reproducing the Double Descent Curve Phenomenon}

In this section, we investigate the double-descent phenomenon in neural networks by systematically varying both the size of the training dataset and the width of a simple feedforward model on the MNIST classification task. Specifically, we use fully connected neural networks with a single hidden layer and ReLU activation. To train them, we use full-batch gradient descent with the AdamW optimizer, without explicit regularization. For each experiment, balanced training subsets of increasing size (100, 1000, 10000, and the full 60000 samples) are constructed to ensure equal class representation, while evaluation is performed on the full MNIST test set. For each fixed training size, the network width is swept across several orders of magnitude (from 2 hidden units to 10000 hidden units), and each model is trained until convergence or zero training error. The final error rates are used to construct log–log plots of error versus network width. 

\begin{figure}
    \centering
\includegraphics[scale=0.5]{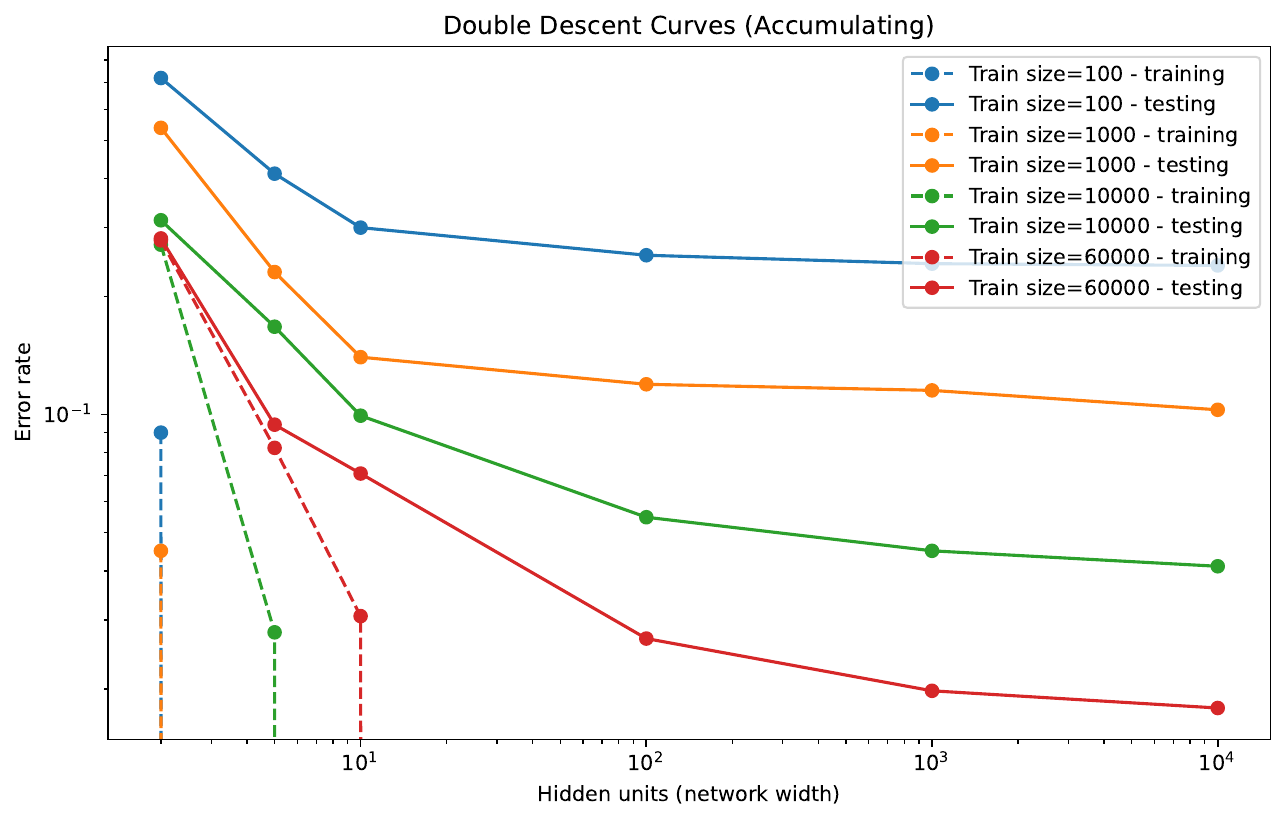}
    \caption{Double descent of test error with respect to network width for varying training set sizes on MNIST.}
    \label{fig:placeholder}
\end{figure}

From the obtained Fig. 1, we observe a partial manifestation, as we explain below, of the double-descent phenomenon across all training set sizes. The x-axis corresponds to the number of hidden neurons (network width), while the y-axis represents the achieved error rate. Each color denotes a different size of the training dataset, with dashed lines indicating training error and solid lines indicating test error. The dashed curves vanish beyond a certain width, indicating that the models achieve zero training error once sufficient capacity is available, marking the onset of interpolation and overfitting. We can see that for all cases considered $10$ (or less) neurons suffice for perfectly interpolating the training datasets, so we do not observe the first descent in this case. In contrast, we immediately observe the second descent in the overfitting regime, where there is negligible training error while the test error decreases monotonically as the number of hidden neurons increases. Moreover, we see that this behavior is consistent across all training set sizes considered. This behavior matches the portion of the double-descent curve that appears beyond the interpolation threshold. Notably, the rate at which the test error decays depends on the amount of training data, with larger datasets exhibiting faster decay and lower overall error levels. This suggests that, in this experimental setting, increased overparameterization does not degrade generalization and instead leads to improved performance, with larger datasets amplifying this effect. It is interesting to note that overfitting helps even in the small-sample (but relatively high-dimensional) regime; it is not limited to situations with plenty of training data.

\section{Proof of Theorem 3.1}

Let function $\mathbf{g}$ be represented by the low-rank tensor product model in (\ref{CPD}),
\begin{equation*}
    \mathbf{g}(\mathbf{x}) = \sum_{r=1}^R \mathbf{v}_r \prod_{n=1}^N g_{n,r}(x_n),
\end{equation*}
where \(g_{n,r} : [0,1] \rightarrow \mathbb{R}\) and \(\mathbf{v}_r \in \mathbb{R}^{M}\). Then, its Dirichlet energy can be expressed in closed form as
\begin{equation}
\begin{split}
\text{DE}(\mathbf{g}) &= \int_{[0,1]^N} \left\|\nabla \mathbf{g}(\mathbf{x})\right\|_F^2 \, d\mathbf{x}\\
&= \int_{[0,1]^N} \left\|\nabla \left(\sum_{r=1}^R \mathbf{v}_r \prod_{n=1}^N g_{n,r}(x_n)\right)\right\|_F^2 \, d\mathbf{x}\\
&= \int_{[0,1]^N} \sum_{m=1}^M\sum_{q=1}^N  \left(\frac{\partial}{\partial x_q}\sum_{r=1}^R \mathbf{v}_r[m] \prod_{n=1}^N g_{n,r}(x_n)\right)^2 \, d\mathbf{x}\\
&= \int_{[0,1]^N} \sum_{r=1}^R \sum_{k=1}^R  \sum_{q=1}^N  \left(\mathbf{v}_r^T \mathbf{v}_k\right) \nabla g_{q,r}(x_q)\nabla q_{q,k}(x_q)\prod_{n\neq q}^N g_{n,r}(x_n)g_{n,k}(x_n) \, d\mathbf{x}\\
&=  \sum_{r=1}^R \sum_{k=1}^R  \sum_{q=1}^N  \left(\mathbf{v}_r^T \mathbf{v}_k\right) \langle\nabla g_{q,r},\nabla q_{q,k}\rangle\prod_{n\neq q}^N \langle g_{n,r}g_{n,k}\rangle\\
&=  \sum_{r=1}^R \sum_{k=1}^R  \sum_{q=1}^N  \left(\mathbf{v}_r^T \mathbf{v}_k\right) \frac{\langle\nabla g_{q,r},\nabla q_{q,k}\rangle}{\langle g_{q,r}g_{q,k}\rangle}\prod_{n=1}^N \langle g_{n,r},g_{n,k}\rangle\\
&=  \sum_{r=1}^R \sum_{k=1}^R    \left(\mathbf{v}_r^T \mathbf{v}_k\right) \left(\prod_{n=1}^N \langle g_{n,r},g_{n,k}\rangle\right)\left(\sum_{q=1}^N\frac{\langle\nabla g_{q,r},\nabla q_{q,k}\rangle}{\langle g_{q,r},g_{q,k}\rangle}\right)\\
&=  \sum_{r=1}^R \sum_{k=1}^R  s_r(\mathbf{g})   \mathbf{A}^{(0)}_{r,k}(\mathbf{g}) \mathbf{A}^{(1)}_{r,k}(\mathbf{g}) s_k(\mathbf{g}) \\
&= \mathbf{s}(\mathbf{g})^\top \mathbf{Z}(\mathbf{g}) \mathbf{s}(\mathbf{g}),    
\end{split}
\end{equation}
where $\mathbf{s}(\mathbf{g}) \in \mathbb{R}^R_+$, $\mathbf{A}^{(0)}(\mathbf{g}) \in \mathbb{R}^{R \times R}$, and $\mathbf{A}^{(1)}(\mathbf{g}) \in \mathbb{R}^{R \times R}$ are defined element-wise as
\begin{equation}
\begin{aligned}
s_r(\mathbf{g}) &= \left\|\mathbf{v}_r\right\| \prod_{n=1}^N \left\|g_{n,r}\right\|,\\
\mathbf{A}^{(0)}_{r,k}(\mathbf{g}) &= \cos(\theta_{\mathbf{v}_r, \mathbf{v}_k}) \prod_{n=1}^N \cos(\theta_{g_{n,r}, g_{n,k}}),\\
\mathbf{A}^{(1)}_{r,k}(\mathbf{g}) &= \sum_{n=1}^N \frac{\langle \nabla g_{n,r}, \nabla g_{n,k} \rangle}{\langle g_{n,r}, g_{n,k} \rangle},\\
\mathbf{Z}_{r,k}(\mathbf{g}) &= A^{(0)}_{r,k}(\mathbf{g}) \, A^{(1)}_{r,k}(\mathbf{g}),
\end{aligned}
\label{Theorem1b}
\end{equation}
and $\langle f, z \rangle := \int_0^1 f(x) z(x) \, dx$ denote the standard $L^2$ inner product, and $\theta_{\cdot, \cdot}$ denotes the angle between two functions in the corresponding inner-product space.

\section{Further results}

\begin{table*}[t]
\centering
\begin{tabular}{lccc}
\toprule
\# neurons/HL \big| Rank & 1000 \big| 11 & 5000 \big| 57 & 20000 \big| 227 \\
\midrule
\multicolumn{4}{l}{\textbf{Best based on Validation Error}} \\
\midrule
NN 1HL wo reg  & 0.941 $\pm$ 0.029 &  0.938 $\pm$ 0.023 &  0.945 $\pm$ 0.032  \\
NN 1HL reg &  0.934  $\pm$ 0.024  & 0.945  $\pm$ 0.024 &  0.930  $\pm$ 0.029  \\
\midrule
NN 2HL wo reg & 0.938 $\pm$ 0.034 & \textbf{0.949 $\pm$ 0.026} & 0.938 $\pm$ 0.019 \\
NN 2HL reg    & 0.949 $\pm$ 0.027 & 0.938 $\pm$ 0.026 & 0.930 $\pm$ 0.032  \\
\midrule
TPBS wo reg     & 0.919 $\pm$ 0.019 & 0.923 $\pm$ 0.027 & 0.894 $\pm$ 0.029 \\
TPBS LDE $\rho$=0.01 & 0.927 $\pm$ 0.019 & 0.912 $\pm$ 0.032 & 0.927 $\pm$ 0.034\\
TPBS LDE $\rho$=0.1  & 0.912 $\pm$ 0.036 & 0.901 $\pm$ 0.036 & \textbf{0.930} $\pm$ \textbf{0.0029}\\
TPBS LDE $\rho$=0.2  & 0.897 $\pm$ 0.029 & 0.901 $\pm$ 0.036 & 0.905 $\pm$ 0.034 \\
\midrule
\multicolumn{4}{l}{\textbf{After Overfitting}} \\
\midrule
NN 1HL wo reg & 0.912 $\pm$ 0.032 & 0.919 $\pm$ 0.034 & 0.916 $\pm$ 0.037  \\
NN 1HL reg    & 0.945 $\pm$ 0.024 & {\color{blue}\textbf{0.952 $\pm$ 0.014}} & 0.945 $\pm$ 0.036 \\
\midrule
NN 2HL wo reg & 0.890 $\pm$ 0.009 & 0.901 $\pm$ 0.018 & 0.897 $\pm$ 0.023  \\
NN 2HL reg    & 0.952 $\pm$ 0.027 & 0.927 $\pm$ 0.029 & 0.919 $\pm$ 0.021  \\
\midrule
TPBS wo reg     & 0.923 $\pm$ 0.024 & 0.919 $\pm$ 0.005 & 0.908 $\pm$ 0.014 \\
TPBS LDE $\rho$=0.01 & 0.923 $\pm$ 0.009 & 0.934 $\pm$ 0.016 & 0.927 $\pm$ 0.026 \\
TPBS LDE $\rho$=0.1  & 0.923 $\pm$ 0.039 & 0.916 $\pm$ 0.041 & 0.930 $\pm$ 0.029 \\
TPBS LDE $\rho$=0.2  & 0.886 $\pm$ 0.023 & 0.919 $\pm$ 0.027 & \textbf{0.938} $\pm$ \textbf{0.010} \\
\bottomrule
\end{tabular}
\caption{Ion Dataset (34D) -- Comparison of 1HL Neural Networks and Tensor Products of B-Splines (TPBS) for a classification task. The reported metric is classification accuracy, shown for best performance of models selected based on validation sets, before and after overfitting. Numbers x \big| y denote the corresponding rank x of a TPBS and the number of neurons y in a neural network per hidden layer. Models appearing in the same column have approximately equal number of parameters, for the appropriate rank or number or neurons. With bold the best per type of model and with blue the best overall. }
\label{tab:Ion}
\end{table*}

\begin{table*}[t]
\centering
\begin{tabular}{lccc}
\toprule
\#neurons/HL \big| Rank & 1000 \big| 11 & 5000 \big| 57 & 20000 \big| 229 \\
\midrule
\multicolumn{4}{l}{\textbf{Best based on Validation Error}} \\
\midrule
NN 1HL wo reg & 0.965 $\pm$ 0.015 & 0.965 $\pm$ 0.018 & 0.967 $\pm$ 0.013 \\
NN 1HL reg    & 0.950 $\pm$ 0.020 & 0.955 $\pm$ 0.027 & 0.957 $\pm$ 0.019  \\
\midrule
NN 2HL wo reg & 0.967 $\pm$ 0.002 & 0.973 $\pm$ 0.008 & {\color{blue}\textbf{0.975 $\pm$ 0.007}} \\
NN 2HL reg    & 0.957 $\pm$ 0.017 & 0.962 $\pm$ 0.018 & 0.957 $\pm$ 0.016  \\
\midrule
TPBS wo reg     & 0.942 $\pm$ 0.034 & 0.948 $\pm$ 0.013 & 0.947 $\pm$ 0.009 \\
TPBS LDE $\rho$=0.01 & 0.950 $\pm$ 0.025 & 0.955 $\pm$ 0.025 & 0.963 $\pm$ 0.016 \\
TPBS LDE $\rho$=0.1  & 0.962 $\pm$ 0.023 & 0.958 $\pm$ 0.012 & \textbf{0.965 $\pm$ 0.014} \\
TPBS LDE $\rho$=0.2  & 0.938 $\pm$ 0.034 & 0.947 $\pm$ 0.030 & 0.952 $\pm$ 0.030 \\
\midrule
\multicolumn{4}{l}{\textbf{After Overfitting}} \\
\midrule
NN 1HL wo reg & 0.952 $\pm$ 0.014 & 0.953 $\pm$ 0.012 & 0.947 $\pm$ 0.018  \\
NN 1HL reg    & 0.950 $\pm$ 0.018 & 0.940 $\pm$ 0.018 & 0.952 $\pm$ 0.019  \\
\midrule
NN 2HL wo reg & 0.952 $\pm$ 0.018 & 0.958 $\pm$ 0.012 & 0.955 $\pm$ 0.011 \\
NN 2HL reg   & 0.953 $\pm$ 0.009 & 0.945 $\pm$ 0.011 & \textbf{0.957 $\pm$ 0.012}  \\
\midrule
TPBS wo reg     & 0.938 $\pm$ 0.036 & 0.940 $\pm$ 0.027 & 0.940 $\pm$ 0.033  \\
TPBS LDE $\rho$=0.01 & 0.960 $\pm$ 0.021 & 0.960 $\pm$ 0.018 & 0.963 $\pm$ 0.016 \\
TPBS LDE $\rho$=0.1  & 0.963 $\pm$ 0.021 & 0.960 $\pm$ 0.012 & \textbf{0.965 $\pm$ 0.014} \\
TPBS LDE $\rho$=0.2  & 0.938 $\pm$ 0.034 & 0.947 $\pm$ 0.030 & 0.952 $\pm$ 0.030 \\
\bottomrule
\end{tabular}
\caption{BCW Dataset (30D) -- Comparison of 1HL Neural Networks and Tensor Products of B-Splines (TPBS) for a classification task. The reported metric is classification accuracy, shown for best performance of models selected based on validation sets, before and after overfitting. Numbers x \big| y denote the corresponding rank x of a TPBS and the number of neurons y in a neural network per one hidden layer. Models appearing in the same column have approximately equal number of parameters, for the appropriate rank or number or neurons. With bold the best per type of model and with blue the best overall.}
\label{tab:BCW}
\end{table*}

\begin{table*}[t]
\centering
\begin{tabular}{lccc}
\toprule
\#neurons/HL \big| Rank & 1000 \big| 12 & 5000 \big| 65  & 20000 \big| 250 \\
\midrule
\multicolumn{4}{l}{\textbf{Best based on Validation Error}} \\
\midrule
NN 1HL wo reg & 0.101 $\pm$ 0.011 & 0.104 $\pm$ 0.015 & 0.104 $\pm$ 0.015  \\
NN 1HL reg  &  0.101 $\pm$ 0.011 & 0.103 $\pm$ 0.014 & 0.103 $\pm$ 0.013 \\
\midrule
NN 2HL wo reg &  0.114 $\pm$ 0.025 & 0.112 $\pm$ 0.026 & 0.104 $\pm$ 0.013 \\
NN 2HL reg  & 0.102 $\pm$ 0.008 & {\color{blue}\textbf{0.101 $\pm$ 0.010}} & 0.102 $\pm$ 0.011  \\
\midrule
TPBS wo reg     & 0.145 $\pm$ 0.013 & 0.139 $\pm$ 0.009  & 0.134 $\pm$ 0.006  \\
TPBS LDE $\rho$=0.01 & 0.154 $\pm$ 0.004 & 0.203 $\pm$ 0.093 & 0.154 $\pm$ 0.033 \\
TPBS LDE $\rho$=0.1  & 0.138 $\pm$ 0.009 & 0.138 $\pm$ 0.009 & \textbf{0.134 $\pm$ 0.005 }\\
TPBS LDE $\rho$=0.2  & 0.140 $\pm$ 0.011 & 0.140 $\pm$ 0.010 & 0.135 $\pm$ 0.007 \\
\midrule
\multicolumn{4}{l}{\textbf{After Overfitting}} \\
\midrule
NN 1HL wo reg & 0.274 $\pm$ 0.019 & 0.231 $\pm$ 0.037 & 0.225 $\pm$ 0.038  \\
NN 1HL reg    & 0.292 $\pm$ 0.021 & 0.231 $\pm$ 0.041 & \textbf{0.223 $\pm$ 0.037}  \\
\midrule
NN 2HL wo reg & 0.494 $\pm$ 0.309 & 0.356 $\pm$ 0.071 & 0.307 $\pm$ 0.046 \\
NN 2HL reg  & 0.428 $\pm$ 0.030 & 0.354 $\pm$ 0.015 & 0.276 $\pm$ 0.050  \\
\midrule
TPBS wo reg     & 0.268 $\pm$ 0.020 & 0.227 $\pm$ 0.003 & 0.227 $\pm$ 0.012 \\
TPBS LDE $\rho$=0.01  & 0.191 $\pm$ 0.030 & 0.238 $\pm$ 0.070 & \textbf{0.173 $\pm$ 0.019} \\
TPBS LDE $\rho$=0.1  & 0.208 $\pm$ 0.012 & 0.228 $\pm$ 0.017 & 0.230 $\pm$ 0.006 \\
TPBS LDE $\rho$=0.2  & 0.295 $\pm$ 0.028 & 0.277 $\pm$ 0.022 & 0.240 $\pm$ 0.005 \\
\bottomrule
\end{tabular}
\caption{Diabetes Dataset (10D) -- Comparison of 1HL Neural Networks and Tensor Products of B-Splines (TPBS) for a regression task. The reported metric is relative mean squared error (RMSE), shown for best performance of models selected based on validation sets, before and after overfitting. Numbers x \big| y denote the corresponding rank x of a TPBS and the number of neurons y in a neural network per hidden layer. Models appearing in the same column have approximately equal number of parameters. With bold the best per type of model and with blue the best overall.}
\label{tab:diabetes}
\end{table*}

\begin{table*}[t]
\centering
\begin{tabular}{lccc}
\toprule
\#neurons/HL \big| Rank & 1000 \big| 14 & 5000 \big| 69 & 20000 \big| 277 \\
\midrule
\multicolumn{4}{l}{\textbf{Best based on Validation Error}} \\
\midrule
NN 1HL wo reg & 0.002 $\pm$ 0.001 & 0.002 $\pm$ 0.001 & 0.003 $\pm$ 0.001 \\
NN 1HL reg    & 0.002 $\pm$ 0.000 & 0.002 $\pm$ 0.001 & 0.003 $\pm$ 0.001 \\
\midrule
NN 2HL wo reg & 0.002 $\pm$ 0.001 & 0.002 $\pm$ 0.001  & 0.002 $\pm$ 0.001  \\
NN 2HL reg & {\color{blue}\textbf{0.001 $\pm$ 0.000}} & 0.001 $\pm$ 0.001 & 0.002 $\pm$ 0.001 \\
\midrule
TPBS wo reg    & 0.002 $\pm$ 0.001 & 0.002 $\pm$ 0.002 & {\color{blue}\textbf{0.001 $\pm$ 0.000}} \\
TPBS LDE $\rho$=0.01 & 0.002 $\pm$ 0.001 & {\color{blue}\textbf{0.001 $\pm$ 0.000}} & {\color{blue}\textbf{0.001 $\pm$ 0.000}} \\
TPBS LDE $\rho$=0.1  & 0.003 $\pm$ 0.003 & {\color{blue}\textbf{0.001 $\pm$ 0.000}} & {\color{blue}\textbf{0.001 $\pm$ 0.000}}\\
TPBS LDE $\rho$=0.2  & {\color{blue}\textbf{0.001 $\pm$ 0.000}} & 0.002 $\pm$ 0.001 & {\color{blue}\textbf{0.001 $\pm$ 0.000}} \\ 
\midrule
\multicolumn{4}{l}{\textbf{After Overfitting}} \\
\midrule
NN 1HL wo reg & 0.002 $\pm$ 0.001 & 0.002 $\pm$ 0.001 & 0.003 $\pm$ 0.001 \\
NN 1HL reg    & 0.002 $\pm$ 0.000 & 0.002 $\pm$ 0.001 & 0.003 $\pm$ 0.001 \\
\midrule
NN 2HL wo reg & {\color{blue}\textbf{0.001 $\pm$ 0.000}} & 0.002 $\pm$ 0.001& 0.002 $\pm$ 0.001  \\
NN 2HL reg & {\color{blue}\textbf{0.001 $\pm$ 0.000}} & 0.001 $\pm$ 0.001 & {\color{blue}\textbf{0.001 $\pm$ 0.000}} \\
\midrule
TPBS wo reg   & 0.002 $\pm$ 0.000 & 0.002 $\pm$ 0.001 & {\color{blue}\textbf{0.001 $\pm$ 0.000}} \\
TPBS LDE $\rho$=0.01 & 0.002 $\pm$ 0.001 & {\color{blue}\textbf{0.001 $\pm$ 0.000}} & {\color{blue}\textbf{0.001 $\pm$ 0.000 }}\\
TPBS LDE $\rho$=0.1  & 0.004 $\pm$ 0.004 & {\color{blue}\textbf{0.001 $\pm$ 0.000}} & {\color{blue}\textbf{0.001 $\pm$ 0.000 }}\\
TPBS LDE $\rho$=0.2  & 0.002 $\pm$ 0.000 & 0.002 $\pm$ 0.001 & {\color{blue}\textbf{0.001 $\pm$ 0.000}}\\
\bottomrule
\end{tabular}
\caption{Yacht Dataset (6D) -- Comparison of 1HL Neural Networks and Tensor Products of B-Splines (TPBS) for a regression task. The reported metric is relative mean squared error (RMSE), shown for best performance  of models selected based on validation sets, before and after overfitting. Numbers x \big| y denote the corresponding rank x of a TPBS and the number of neurons y in a neural network per hidden layer. Models appearing in the same column have approximately equal number of parameters. With bold the best per type of model and with blue the best overall.}
\label{tab:yacht}
\end{table*}

\begin{table*}[t]
\centering
\begin{tabular}{lccccc}
\toprule
\#neurons/HL \big| Rank & 1000 \big| 13 & 5000 \big| 64 & 20000 \big| 254 \\
\midrule
\multicolumn{4}{l}{\textbf{Best based on Validation}} \\
\midrule
NN 1HL wo      & $0.265 \pm 0.008$ & $0.265 \pm 0.008$ & $0.266 \pm 0.007$ \\
NN 1HL reg     & $0.264 \pm 0.008$ & $0.264 \pm 0.007$ & $0.265 \pm 0.007$ \\
\midrule
NN 2HL wo      & $0.263 \pm 0.014$ & $0.264 \pm 0.008$ & ${\color{blue}\bm{0.263 \pm 0.010}}$ \\
NN 2HL reg     & $0.269 \pm 0.011$ & $0.265 \pm 0.009$ & $0.270 \pm 0.005$ \\
\midrule
TPBS wo        & $0.350 \pm 0.006$ & $0.348 \pm 0.005$ & $0.444 \pm 0.024$ \\
TPBS LDE $\rho$=0.01 & $0.355 \pm 0.007$ & $\bm{0.340 \pm 0.012}$ & $0.433 \pm 0.036$ \\
TPBS LDE $\rho$=0.1  & $0.350 \pm 0.005$ & $0.348 \pm 0.004$ & $0.464 \pm 0.008$ \\
TPBS LDE $\rho$=0.2  & $0.349 \pm 0.005$ & $0.348 \pm 0.005$ & $0.448 \pm 0.020$ \\
\midrule
\multicolumn{4}{l}{\textbf{After Overfitting}} \\
\midrule
NN 1HL wo reg & 0.613 $\pm$ 0.038 & 0.560 $\pm$ 0.039 & \textbf{0.533} $\pm$ \textbf{0.075}  \\
NN 1HL reg    & 0.649 $\pm$ 0.158 & 0.545 $\pm$ 0.046 & 0.536 $\pm$ 0.069\\
\midrule
NN 2HL wo reg &  0.756 $\pm$ 0.132 & 0.689 $\pm$ 0.041 & 0.730 $\pm$ 0.105 \\
NN 2HL reg    & 0.851 $\pm$ 0.123 & 0.760 $\pm$ 0.105 & 0.654 $\pm$ 0.074 \\
\midrule
TPBS wo reg      & 0.755 $\pm$ 0.041 & 0.965 $\pm$ 0.055  & 1.314 $\pm$ 0.271 \\
TPBS LDE $\rho$=0.01  & 0.376 $\pm$ 0.025 & $\bm{0.336 \pm 0.013}$ & 0.430 $\pm$ 0.032 \\
TPBS LDE $\rho$=0.1   & 0.552 $\pm$ 0.037 & 0.393 $\pm$ 0.018 & 0.478 $\pm$ 0.030 \\
TPBS LDE $\rho$=0.2   & 0.576 $\pm$ 0.019 & 0.508 $\pm$ 0.021 & 0.510 $\pm$ 0.007 \\
\bottomrule
\end{tabular}
\caption{Physico Dataset (9D) -- Comparison of 1HL and 2HL Neural Networks and Tensor Products of B-Splines (TPBS) for a regression task. The reported metric is relative mean squared error (RMSE), shown for best performance  of models selected based on validation sets, before and after overfitting. Numbers x \big| y denote the corresponding rank x of a TPBS and the number of neurons y in a neural network per hidden layer. Models appearing in the same column have approximately equal number of parameters. }
\label{tab:physico}
\end{table*}

\begin{table*}[t]
\centering
\begin{tabular}{lccccc}
\toprule
\#neurons/HL \big| Rank & 1000 \big| 11 & 5000 \big| 57 & 20000 \big| 228  \\
\midrule
\multicolumn{4}{l}{\textbf{Best Validation}} \\
\midrule
NN 1HL wo reg & 0.054 $\pm$ 0.004 & 0.049 $\pm$ 0.002 & {\color{blue}\textbf{0.049} $\pm$ \textbf{0.001}} \\
NN 1HL reg   & 0.053 $\pm$ 0.004 & 0.050 $\pm$ 0.003 & 0.049 $\pm$ 0.002 \\
\midrule
NN 2HL wo reg & 0.058 $\pm$ 0.006 & 0.062 $\pm$ 0.006 & 0.061 $\pm$ 0.005 \\
NN 2HL reg    & 0.059 $\pm$ 0.004 & 0.057 $\pm$ 0.007 & 0.057 $\pm$ 0.005 \\
\midrule
TPBS wo  reg   & $0.457 \pm 0.003$ & $0.437 \pm 0.012$ & $0.445 \pm 0.019$ \\
TPBS LDE $\rho$=0.01 & $0.356 \pm 0.070$ & \bm{$0.240 \pm 0.042$} & $0.249 \pm 0.021$ \\
TPBS LDE $\rho$=0.1  & $0.461 \pm 0.003$ & $0.329 \pm 0.016$ & $0.414 \pm 0.030$ \\
TPBS LDE $\rho$=0.2  & $0.461 \pm 0.002$ & $0.383 \pm 0.015$ & $0.443 \pm 0.018$ \\
\midrule
\multicolumn{4}{l}{\textbf{After Overfitting}} \\
\midrule
NN 1HL wo reg & 0.094 $\pm$ 0.006 & 0.061 $\pm$ 0.004 & \textbf{0.060} $\pm$ \textbf{0.005}  \\
NN 1HL reg    & 0.075 $\pm$ 0.006 & 0.063 $\pm$ 0.005 & 0.061 $\pm$ 0.005 \\
\midrule
NN 2HL wo reg & 0.103 $\pm$ 0.008 & 0.098 $\pm$ 0.009 & 0.076 $\pm$ 0.002 \\
NN 2HL reg    & 0.106 $\pm$ 0.031 & 0.092 $\pm$ 0.021  & 0.079 $\pm$ 0.007 \\
\midrule
TPBS wo reg      & 1.937 $\pm$ 0.582 & 0.444 $\pm$ 0.014 & 0.476 $\pm$ 0.016 \\
TPBS LDE $\rho$=0.01  & 0.538 $\pm$ 0.326 & \textbf{0.243} $\pm$ \textbf{0.046} & 0.251 $\pm$ 0.022 \\
TPBS LDE $\rho$=0.1   & 0.820 $\pm$ 0.145 & 0.329 $\pm$ 0.016 & 0.413 $\pm$ 0.030 \\
TPBS LDE $\rho$=0.2   & 1.051 $\pm$ 0.378 & 0.383 $\pm$ 0.015 & 0.476 $\pm$ 0.015 \\
\bottomrule
\end{tabular}
\caption{Sarcos Dataset (21D) -- Comparison of 1HL and 2HL Neural Networks and Tensor Products of B-Splines (TPBS) for a regression task. The reported metric is relative mean squared error (RMSE), shown for best performance  of models selected based on validation sets, before and after overfitting. Numbers x \big| y denote the corresponding rank x of a TPBS and number of neurons y in a neural network per hidden layer. Models appearing in the same column have approximately equal number of parameters. With bold the best per type of model and with blue the best overall.}
\label{tab:sarcos}
\end{table*}

\end{document}